\pdfoutput=1
\documentclass[journal]{IEEEtran}

\usepackage[T1]{fontenc}
\usepackage{cite}
\usepackage{amsmath,amssymb}
\usepackage{graphicx}
\usepackage{booktabs}
\usepackage{multirow}
\usepackage{array}
\usepackage{subcaption}
\usepackage{url}
\usepackage[hidelinks]{hyperref}
\usepackage{balance}

\title{ROSA-TFormer: A Radar-Optical Sensor-Aware Temporal Transformer for \textit{Pinus sylvestris} Plantation Classification in Northern Shaanxi Using GEE-Derived Sentinel-1/2 Time Series}

\author{
Nengbo Zhang, Sheng Chang.  
\thanks{Nengbo Zhang and  Sheng Chang is with the Key Laboratory of Remote Sensing and Digital Earth, Aerospace Information Research Institute, Chinese Academy of Sciences (AIRCAS), Beijing 100101, China. E-mail: bookerznb@mail.com.}
}

\begin{document}
\maketitle

\begin{abstract}
Mapping \textit{Pinus sylvestris} var. \textit{mongolica} plantations is important for evaluating shelterbelt restoration in semi-arid northern Shaanxi, where pine stands occur in fragmented mosaics of bare land, built-up areas, other vegetation, and water bodies. We propose ROSA-TFormer, a Radar-Optical Sensor-Aware Temporal Transformer that classifies \textit{P. sylvestris} from Google Earth Engine (GEE)-composited Sentinel-1/2 time series. The model separates SAR and optical embeddings, uses token-level sensor gating, and combines a learnable class token with attention-based temporal pooling to represent radar-optical phenology. We evaluated two point-level datasets from the study area, a 12-token monthly set and a 24-token half-month set. On HalfMonth-dataBig, ROSA-TFormer achieved 99.67\% overall accuracy, 99.56\% macro F1, and 98.91\% \textit{P. sylvestris} F1 under strict point-level validation, with three-seed averages of 99.61$\pm$0.17\% overall accuracy and 99.49$\pm$0.26\% macro F1. Under a deterministic coordinate-based spatial split, it achieved 99.34\% overall accuracy and 99.08\% macro F1, improving over the early-fusion Transformer but not the strongest 1D-CNN baseline. Removing the sensor gate reduced macro F1 from 99.56\% to 99.42\%. These results show that GEE-based radar-optical temporal sampling and sensor-aware attention provide strong point-level evidence for \textit{P. sylvestris} classification, while wall-to-wall validation remains needed.
\end{abstract}

\begin{IEEEkeywords}
\textit{Pinus sylvestris} var. \textit{mongolica}, northern Shaanxi, Google Earth Engine, Sentinel-1, Sentinel-2, radar-optical fusion, sensor-aware fusion, temporal Transformer, Three-North Shelter Forest Program, artificial plantation forest.
\end{IEEEkeywords}

\section{Introduction}
\IEEEPARstart{P}{inus} sylvestris var. \textit{mongolica} Litv., commonly known in China as Zhangzisong, is one of the most important plantation conifers used in windbreak and sand-fixation programs in northern China \cite{zheng2012,zhai2023}. Northern Shaanxi is a typical semi-arid shelterbelt and ecological restoration region within the broader Three-North Shelter Forest context. Since the launch of the Three-North Shelter Forest Program in 1978, large areas of shelterbelt plantations have been established across the northwest, north, and northeast regions of China to reduce wind erosion, slow desertification, protect agricultural production, and improve ecological resilience \cite{zhai2023}. As ecological restoration programs increasingly require quantitative assessment of plantation survival, distribution, and carbon-related functions, reliable mapping of \textit{P. sylvestris} plantations in northern Shaanxi has become a necessary remote sensing task.

The classification of \textit{P. sylvestris} in northern Shaanxi shelterbelt landscapes is difficult for three reasons. First, the target species is often embedded in fragmented semi-arid mosaics, where plantations are adjacent to bare land, rural construction land, cropland, shrubland, and water bodies \cite{zheng2012,li2024trees}. A classifier trained only for generic forest/non-forest separation is therefore insufficient. Second, young or sparse \textit{P. sylvestris} stands may contain exposed soil and understory signals, producing spectral confusion with bare land and other vegetation. Third, optical imagery alone is vulnerable to cloud contamination and seasonal spectral convergence, particularly when deciduous vegetation reaches peak greenness during the growing season \cite{hemmerling2021,li2024trees}.

\begin{figure*}[!t]
\centering
\includegraphics[width=\textwidth]{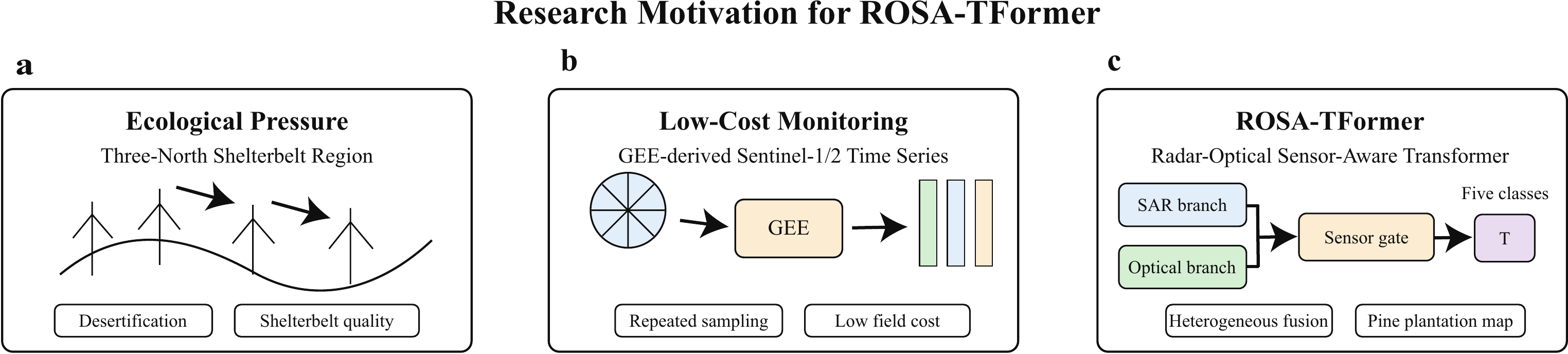}
\caption{Research motivation of GEE-based radar-optical time series learning for \textit{P. sylvestris} plantation monitoring. Ecological pressure and sparse field-survey efficiency create a need for scalable observation; GEE-derived Sentinel-1/2 sequences provide low-cost repeated monitoring data; ROSA-TFormer then learns radar-optical temporal features for accurate pine plantation mapping.}
\label{fig:research_motivation}
\end{figure*}

Multi-source satellite time series provide an effective way to address these limitations. Sentinel-2 multispectral imagery captures vegetation pigment, water content, red-edge response, and shortwave infrared reflectance, while Sentinel-1 C-band synthetic aperture radar (SAR) provides structural and moisture-related information that is less affected by cloud cover and solar illumination \cite{torres2012,drusch2012}. In addition, the annual temporal trajectory contains species-relevant phenological information. Evergreen conifers tend to maintain more stable winter reflectance and radar volume scattering than deciduous vegetation. These properties suggest that \textit{P. sylvestris} classification should be treated as a multivariate radar-optical time series problem rather than only a single-date spectral classification problem. Fig.~\ref{fig:research_motivation} summarizes the ecological motivation, the need for scalable GEE-based monitoring, and the role of ROSA-TFormer in pine plantation mapping.

Traditional machine learning methods such as Random Forest (RF) and XGBoost have been widely used in remote sensing classification because they are robust, easy to train, and effective for high-dimensional feature spaces \cite{breiman2001,belgiu2016,chen2016}. However, these methods usually depend on manually engineered descriptors, including band statistics, vegetation indices, and flattened temporal features. Such descriptors provide useful summary information but may not fully capture non-local seasonal dependencies or sensor-specific contributions across the annual sequence. Deep learning approaches, especially one-dimensional convolutional neural networks and attention-based models, can directly learn discriminative temporal representations from remote sensing sequences \cite{zhong2019,russwurm2020,vaswani2017}. Among them, Transformers are particularly attractive because self-attention can relate observations from different months without the recurrence constraints of long short-term memory networks or the locality assumptions of convolution.

These requirements expose three methodological gaps. First, tabular classifiers such as RF and XGBoost rely on manually summarized temporal descriptors and may lose non-local seasonal relationships. Second, generic one-dimensional CNN or Transformer models usually treat the annual Sentinel-1/2 sequence as a single fused channel stream, without explicitly modeling the different noise sources and physical meanings of SAR and optical observations. Third, the effect of monthly versus denser half-month temporal sampling remains unclear when the task is evaluated under strict point-level validation. Motivated by these gaps, this study proposes ROSA-TFormer, a Radar-Optical Sensor-Aware Temporal Transformer for \textit{P. sylvestris} classification using Sentinel-1/2 monthly and half-month time series. The main contributions are as follows.

\begin{enumerate}
\item We formulate \textit{P. sylvestris} identification as an application-driven artificial plantation forest classification problem in northern Shaanxi, explicitly considering confusing background classes common in semi-arid afforestation landscapes: bare land, built-up area, other vegetation, and water body.
\item We construct and compare GEE-derived Sentinel-1/2 point time-series datasets at 12-token monthly and 24-token half-month resolutions, and further conduct a same-sample 12/24-token comparison to separate temporal granularity from sample-scale effects.
\item We propose ROSA-TFormer, a Radar-Optical Sensor-Aware Temporal Transformer with separate SAR/optical tokenization, adaptive sensor gating, learnable class-token aggregation, and temporal attention pooling. This architecture is designed specifically for heterogeneous Sentinel-1/2 annual sequences rather than directly reusing a generic Transformer encoder.
\item We strengthen the experimental protocol by reconstructing GEE records into point-level annual sequences and evaluating the model with matched baselines, module ablations, deterministic spatial splitting, multi-seed statistics, and gate/attention interpretation.
\end{enumerate}

\section{Related Work}
\subsection{Tree Species Classification From Satellite Time Series}
Tree species mapping has long benefited from multi-temporal optical imagery, because different species exhibit distinct phenological trajectories in greenness, red-edge reflectance, canopy water content, and senescence timing \cite{gomez2016,belgiu2018}. Sentinel-2 has become a major data source for vegetation and tree-species classification because of its 10--20 m spatial resolution, red-edge bands, and frequent revisit cycle \cite{drusch2012}. Studies on forest type mapping have shown that dense time series often outperform single-date images, especially when the species of interest are phenologically distinct.

For \textit{P. sylvestris} and related coniferous plantations, the remote sensing challenge is not only spectral separability but also ecological context. In northern Shaanxi, plantations may be sparse, mixed with bare soil, or distributed in small patches along shelterbelt corridors. Previous work on coniferous forest mapping in semi-arid regions has demonstrated the value of combining optical and radar information, but many studies still focus on broad coniferous categories rather than a specific plantation species. The present work addresses this gap by targeting \textit{P. sylvestris} in a northern Shaanxi shelterbelt-oriented classification setting.

\subsection{Sentinel-1/2 Data Fusion}
Sentinel-1 and Sentinel-2 provide complementary information for land cover mapping. Sentinel-2 optical bands are sensitive to chlorophyll, leaf structure, canopy water content, and soil background, whereas Sentinel-1 SAR backscatter is sensitive to surface roughness, canopy structure, dielectric properties, and moisture \cite{torres2012,dostalova2018}. Fusion of these sensors has improved classification in agricultural, forest, wetland, and urban contexts \cite{clerici2017,tavares2019}. In semi-arid shelterbelt landscapes, this complementarity is especially important because spectral differences between sparse vegetation and bare soil can be weak during dry periods, while SAR backscatter may preserve useful structural contrasts.

Temporal fusion further improves separability. For example, water bodies are characterized by persistently low SAR backscatter, built-up areas by high and stable backscatter, deciduous vegetation by strong seasonal optical changes, and evergreen conifers by relatively stable winter signals. The proposed method uses this full annual behavior rather than relying on a small number of selected dates.

\subsection{Deep Learning and Transformer Models in Remote Sensing}
Deep learning has been increasingly applied to remote sensing time series classification. One-dimensional CNNs can extract local temporal patterns and have been successful in crop and vegetation classification \cite{zhong2019}. Recurrent neural networks and LSTMs can model sequential dependencies, but their training can be slow and their ability to capture long-range interactions may be limited. Transformers replace recurrence with self-attention and have achieved strong performance in natural language processing and time series modeling \cite{vaswani2017}. In remote sensing, self-attention has been used for raw satellite time series classification and has shown promise for modeling complex temporal relationships \cite{russwurm2020,russwurm2019,yuan2023}.

For \textit{P. sylvestris} mapping, a Transformer is conceptually suitable because discriminative evidence may be distributed across non-adjacent time periods. Winter stability, spring transition, summer peak reflectance, and autumn decline jointly define the annual signature. Self-attention allows the model to compare these time steps directly and learn which temporal relationships are most relevant for classification.

\section{Study Area and Data}
\subsection{Study Area}
The study uses point samples from artificial plantation forest sites in the northern Shaanxi shelterbelt region and adjacent semi-arid landscapes. According to the coordinates embedded in the GEE-exported CSV files, the available samples span approximately 108.27$^\circ$--111.92$^\circ$E and 37.90$^\circ$--39.47$^\circ$N. This coordinate range is consistent with a northern Shaanxi and Ordos-Plateau-edge afforestation context rather than an urban Beijing-only study area. The local project directory name is therefore treated only as a data-management label and is not used as the geographic definition of the experiment.

The sampled landscapes contain built-up areas, water bodies, cropland or other vegetation, bare or sparsely vegetated land, and artificial coniferous plantations. Seasonal precipitation and soil-moisture variation create strong temporal changes in vegetation and background reflectance. \textit{P. sylvestris} plantations are commonly distributed in hilly or sandy shelterbelt environments as ecological restoration stands. Their evergreen canopy and drought tolerance make them an important target for ecological monitoring in northern Shaanxi. The experiments in this paper should therefore be interpreted as a northern Shaanxi point-sample case study motivated by the broader Three-North artificial plantation forest context, rather than as a completed wall-to-wall assessment of the full Three-North region. GEE is used because it can consistently extract Sentinel observations for spatially dispersed afforestation sites and can support larger-area extensions when wall-to-wall inference data and independent validation samples become available.

\begin{figure*}[!t]
\centering
\includegraphics[width=\textwidth]{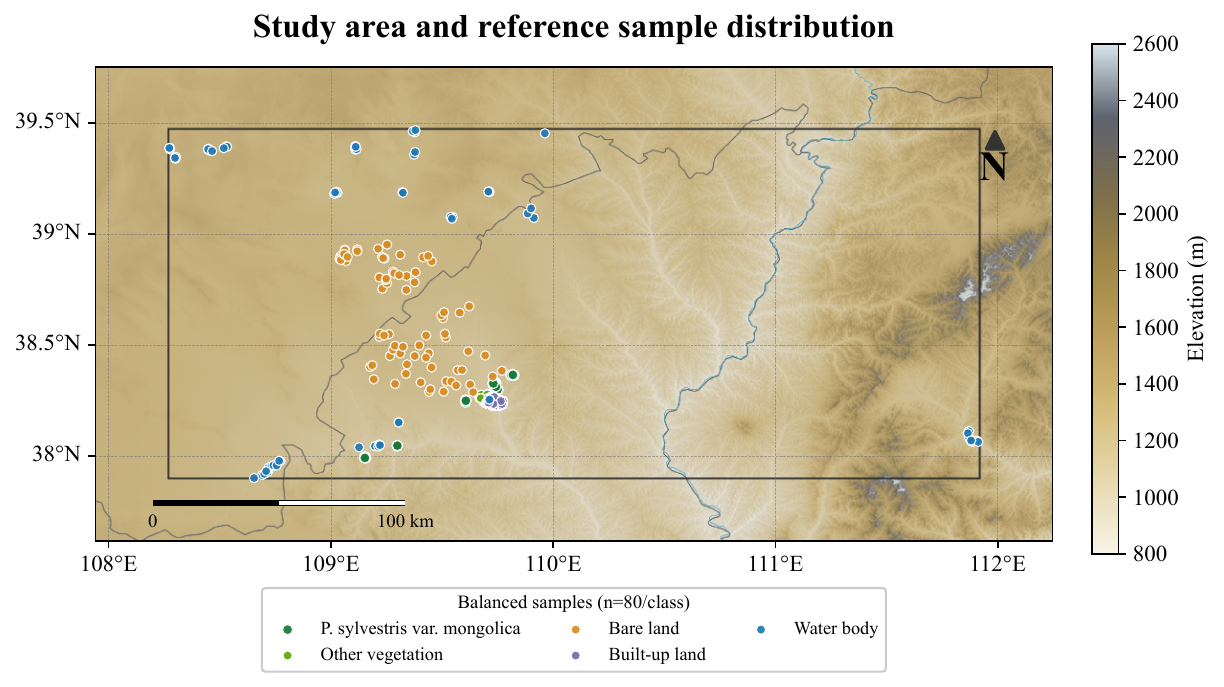}
\caption{Research area and GEE reference sample distribution in the northern Shaanxi shelterbelt region. The map was produced with GEE and geemap/cartoee using SRTM90 elevation and hillshade, HydroSHEDS river features, China province boundaries, and a spatially balanced display of 80 points per class from HalfMonth-dataBig. The full reference set contains 3989 deduplicated point sequences.}
\label{fig:study_area}
\end{figure*}

Fig.~\ref{fig:study_area} shows the northern Shaanxi study area and the actual GEE reference points used in the experiment. The map combines an elevation and hillshade background, province-level boundary context, HydroSHEDS river features, and a balanced display of the five reference classes to clarify both terrain setting and sample distribution.

\subsection{Reference Classes}
\begin{figure*}[!t]
\centering
\includegraphics[width=0.92\textwidth]{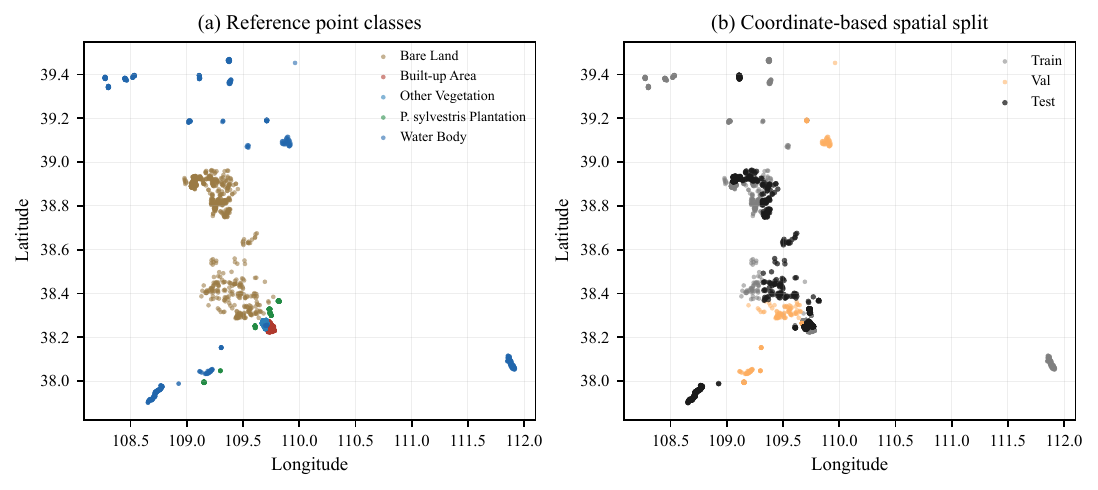}
\caption{Spatial distribution of HalfMonth-dataBig reference points and the deterministic coordinate-based spatial split. The coordinate range indicates that the available samples represent a northern Shaanxi shelterbelt case-study setting rather than a Beijing-only study area.}
\label{fig:sample_distribution}
\end{figure*}

Five land-cover classes were used: bare land, built-up area, other vegetation, water body, and \textit{P. sylvestris} plantation, locally known as Zhangzisong. The reference points were collected for artificial plantation forest interpretation, with emphasis on \textit{P. sylvestris} stands and their surrounding land-cover background in northern Shaanxi and adjacent shelterbelt landscapes. The inclusion of bare land and built-up area is not incidental. In this context, these classes are frequent sources of confusion because afforestation sites often occur near exposed soil, rural settlements, construction land, roads, and agricultural margins. Table~\ref{tab:dataset} summarizes the sample distribution, and Fig.~\ref{fig:sample_distribution} shows the point locations and spatial split used for generalization testing.

\begin{table}[!t]
\centering
\caption{Two GEE-derived datasets used for strict point-level validation. Monthly-data contains 12-token annual sequences, while HalfMonth-dataBig contains 24-token annual sequences.}
\label{tab:dataset}
\begin{tabular}{lrrrr}
\toprule
Dataset & Records & Seq. & Len. & Test \\
\midrule
Monthly-data & 10157 & 522 & 12 & 157 \\
HalfMonth-dataBig & 73733 & 3989 & 24 & 1197 \\
\bottomrule
\end{tabular}
\end{table}

\begin{table*}[!t]
\centering
\caption{Class-wise point-sequence counts for the two strict point-level train/validation/test splits.}
\label{tab:class_counts}
\begin{tabular}{llrrrrrr}
\toprule
Dataset & Split & Bare land & Built-up & Other veg. & Water & \textit{P. sylvestris} & Total \\
\midrule
\multirow{3}{*}{Monthly-data} & Train & 62 & 69 & 77 & 67 & 35 & 310 \\
& Val. & 11 & 12 & 14 & 12 & 6 & 55 \\
& Test & 32 & 35 & 39 & 34 & 17 & 157 \\
\midrule
\multirow{3}{*}{HalfMonth-dataBig} & Train & 548 & 540 & 545 & 557 & 183 & 2373 \\
& Val. & 97 & 95 & 96 & 98 & 33 & 419 \\
& Test & 276 & 272 & 275 & 281 & 93 & 1197 \\
\bottomrule
\end{tabular}
\end{table*}

\subsection{GEE-Based Sentinel-1 and Sentinel-2 Time Series}
All remote sensing samples were extracted through Google Earth Engine, which provides cloud-based access to analysis-ready satellite archives and scalable geospatial computation \cite{gorelick2017}. GEE was used to query Sentinel-1 and Sentinel-2 imagery, filter observations by date and region, generate monthly or half-month composite windows, and export point-level time series for subsequent model training. This design is important for the present task because the artificial plantation forest sites in northern Shaanxi are spatially dispersed, and repeated local downloading of all Sentinel scenes would be inefficient.

Sentinel-1 Ground Range Detected products in Interferometric Wide Swath mode were used to obtain VV and VH C-band SAR backscatter. Sentinel-2 Level-2A products were used for ten optical bands: B2, B3, B4, B5, B6, B7, B8, B8A, B11, and B12. These bands cover visible, red-edge, near-infrared, and shortwave infrared spectral regions, providing sensitivity to vegetation vigor, canopy structure, and water content.

For the monthly dataset, records from the same month were averaged to form a 12-token annual sequence. For the half-month dataset, observations were composited into a full 24-window calendar covering 2024. Missing temporal positions were filled through linear interpolation followed by forward/backward filling when necessary. Each sample was therefore represented as a sequence:
\begin{equation}
\mathbf{X}_i = [\mathbf{x}_{i,1},\mathbf{x}_{i,2},\ldots,\mathbf{x}_{i,T}],\quad \mathbf{x}_{i,t}\in\mathbb{R}^{12},
\end{equation}
where $T=12$ for the monthly dataset and $T=24$ for the half-month dataset. The 12 channels include two SAR bands and ten optical bands.

\section{Methodology}
\subsection{Overall Framework}
\begin{figure*}[!t]
\centering
\includegraphics[width=0.98\textwidth]{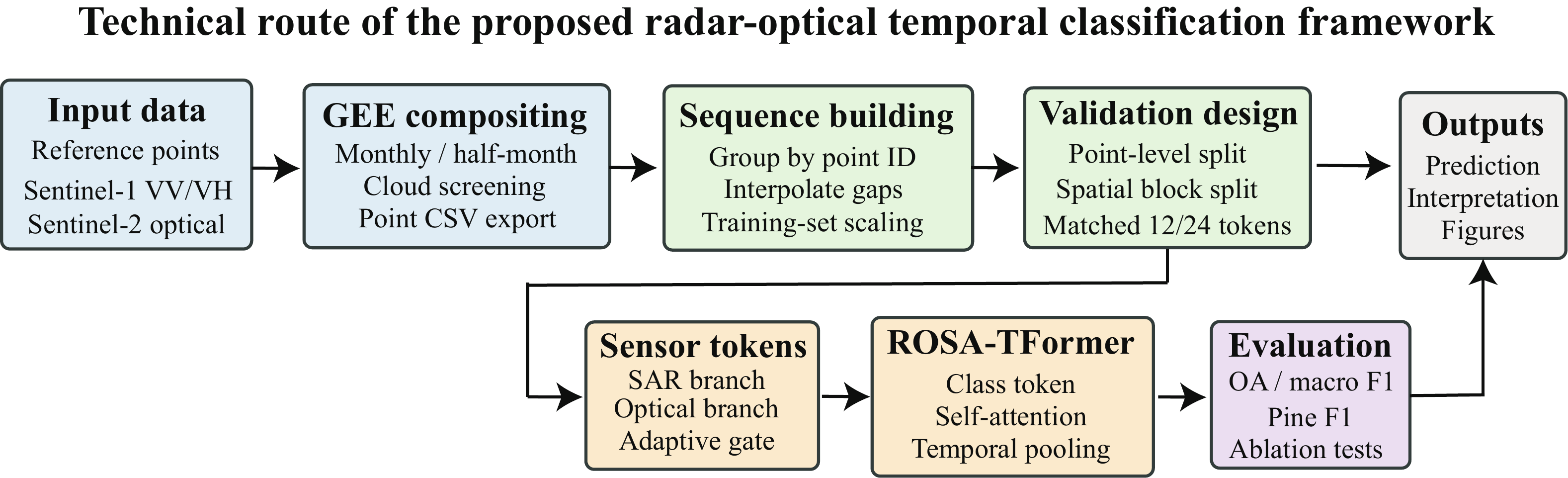}
\caption{Technical route of the proposed radar-optical temporal classification framework. The workflow starts from reference points and Sentinel-1/2 observations, constructs leakage-safe point-level sequences through GEE compositing and interpolation, trains ROSA-TFormer with sensor-aware temporal tokens, and evaluates the resulting model through point-level, spatial, temporal-granularity, and ablation experiments.}
\label{fig:technical_route}
\end{figure*}

The proposed framework contains four stages. First, Sentinel-1 and Sentinel-2 observations are queried and composited in GEE into monthly or half-month sequences over artificial plantation forest sample sites. Second, point-level sequences are built by grouping records using class labels and point identifiers. Third, sequences are standardized using training-set statistics to avoid test-set information leakage. Fourth, ROSA-TFormer learns a class probability distribution for each sample through radar-optical sensor-aware temporal attention. Fig.~\ref{fig:technical_route} summarizes the complete technical route from reference data and GEE compositing to sequence construction, validation, sensor-aware temporal learning, and interpretation outputs.

\subsection{Sequence Reconstruction and Normalization}
The implementation reconstructs one complete annual sequence for each unique pair of class label and point identifier. If multiple records exist for the same point and temporal interval, their band values are averaged before sequence construction. The time axis is then reindexed to the complete set of monthly or half-month windows. Missing temporal observations are filled using linear interpolation with bidirectional limits, followed by forward and backward filling. Remaining missing values, which occur only when interpolation cannot recover a valid value, are set to zero. This procedure converts irregular GEE-exported point observations into a dense tensor $\mathbf{X}\in\mathbb{R}^{N\times T\times12}$, where $T=12$ or $T=24$ depending on the dataset.

For deep learning models, normalization is performed channel-wise using only the training subset:
\begin{equation}
\hat{x}_{i,t,c}=\frac{x_{i,t,c}-\mu_c^{\text{train}}}{\sigma_c^{\text{train}}+\epsilon},
\end{equation}
where $\mu_c^{\text{train}}$ and $\sigma_c^{\text{train}}$ are computed by flattening all training samples and time steps for channel $c$, and $\epsilon=10^{-8}$ prevents numerical instability. The test set is transformed using the same training statistics, preventing information leakage.

\subsection{ROSA-TFormer Architecture}
\begin{figure*}[!t]
\centering
\includegraphics[width=\textwidth]{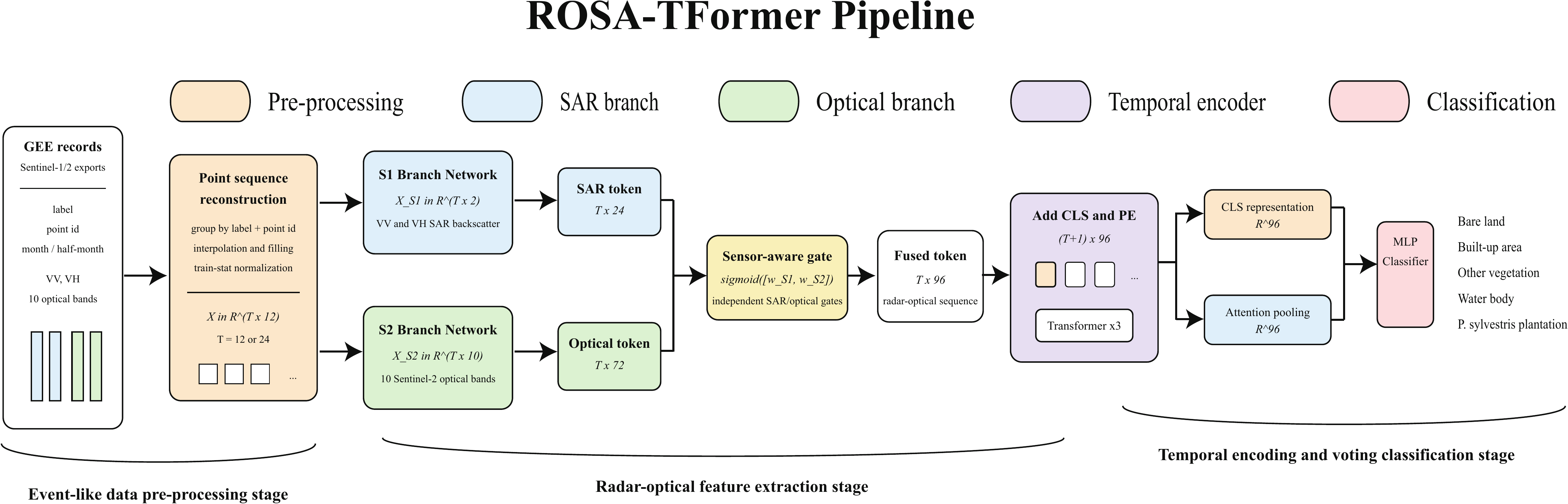}
\caption{Overall architecture of the proposed ROSA-TFormer, redrawn by the author as an Adobe Illustrator vector pipeline. Sentinel-1 SAR and Sentinel-2 optical time series are first cleaned and reconstructed from GEE-derived point observations. Separate radar and optical branches embed heterogeneous sensor channels, a radar-optical sensor-aware gate adaptively fuses the two embeddings, temporal encoding models annual dependencies, and dual aggregation combines the encoded class-token representation with an attention-pooled temporal representation before MLP classification.}
\label{fig:rosa_architecture}
\end{figure*}

The proposed ROSA-TFormer treats a remote sensing sample as a one-dimensional multivariate temporal sequence. Unlike image Transformers that divide an image into spatial patches, the proposed network operates along the temporal axis. Each monthly or half-month observation is a token containing heterogeneous Sentinel-1 SAR and Sentinel-2 optical channels. Compared with the initial plain temporal Transformer, ROSA-TFormer introduces four modifications: sensor-specific embedding branches, token-level SAR/optical gating, learnable class-token aggregation, and attention-based temporal pooling. Fig.~\ref{fig:rosa_architecture} illustrates the overall data flow from GEE-derived radar-optical sequences to the final five-class prediction.

\subsubsection{Sensor-Aware Tokenization}
The raw 12-dimensional observation at time $t$ is divided into a two-channel SAR vector $\mathbf{x}_{t}^{S1}$ and a ten-channel optical vector $\mathbf{x}_{t}^{S2}$. They are embedded by separate projection branches:
\begin{equation}
\mathbf{e}_{t}^{S1}=g_1(\mathbf{W}_{1}\mathbf{x}_{t}^{S1}+\mathbf{b}_{1}),\quad
\mathbf{e}_{t}^{S2}=g_2(\mathbf{W}_{2}\mathbf{x}_{t}^{S2}+\mathbf{b}_{2}),
\end{equation}
where $g_1(\cdot)$ and $g_2(\cdot)$ denote layer normalization and GELU activation. In the optimized implementation, the SAR branch produces 24 dimensions, the optical branch produces 72 dimensions, and the concatenated token has $d=96$ dimensions.

Because SAR and optical observations contribute differently across seasons and land-cover types, ROSA-TFormer further learns a token-level sensor gate:
\begin{equation}
\boldsymbol{\alpha}_t=\sigma(\mathbf{W}_g[\mathbf{e}_{t}^{S1};\mathbf{e}_{t}^{S2}]+\mathbf{b}_g),
\end{equation}
where $\boldsymbol{\alpha}_t\in\mathbb{R}^{2}$ weights the SAR and optical branches. The two gate values are independent sigmoid responses rather than softmax-normalized sensor proportions, so $\alpha_{t,1}+\alpha_{t,2}$ is not constrained to one. The final temporal token is
\begin{equation}
\mathbf{z}_t=[\alpha_{t,1}\mathbf{e}_{t}^{S1};\alpha_{t,2}\mathbf{e}_{t}^{S2}].
\end{equation}

\subsubsection{Temporal Positional Encoding}
Because self-attention is permutation-invariant, temporal order is injected using learnable positional embeddings. ROSA-TFormer also prepends a learnable class token $\mathbf{z}_{cls}$ to the temporal tokens:
\begin{equation}
\mathbf{H}^{0}=[\mathbf{z}_{cls};\mathbf{z}_1,\ldots,\mathbf{z}_{T}]+\mathbf{P}.
\end{equation}

\subsubsection{Self-Attention Encoder}
The encoder contains three Transformer layers with four attention heads. For each head, queries, keys, and values are computed from the input sequence:
\begin{equation}
\mathbf{Q}=\mathbf{H}\mathbf{W}_Q,\quad
\mathbf{K}=\mathbf{H}\mathbf{W}_K,\quad
\mathbf{V}=\mathbf{H}\mathbf{W}_V.
\end{equation}
Scaled dot-product attention is then defined as:
\begin{equation}
\text{Attention}(\mathbf{Q},\mathbf{K},\mathbf{V})
=\text{softmax}\left(\frac{\mathbf{Q}\mathbf{K}^{T}}{\sqrt{d_k}}\right)\mathbf{V}.
\end{equation}
This operation enables the model to compare observations across the full annual sequence, for example linking winter SAR stability with summer NIR response.

\subsubsection{Dual Aggregation and Classification}
\begin{figure}[!t]
\centering
\includegraphics[width=0.98\linewidth]{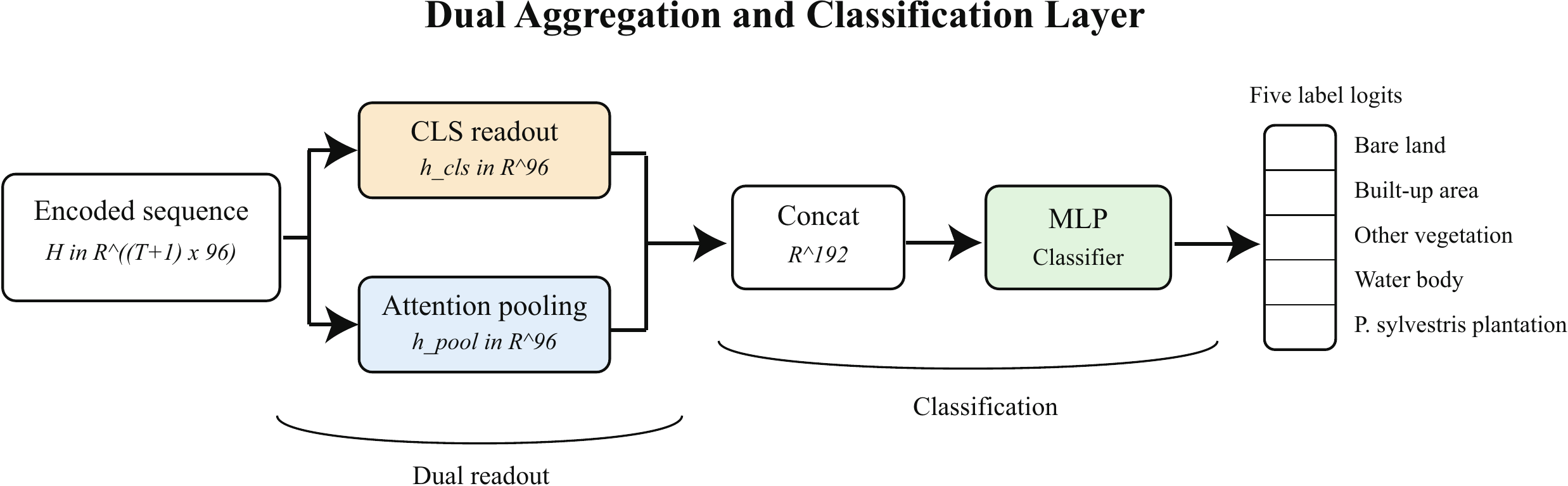}
\caption{Simplified dual aggregation and classification layer. The encoded sequence is read through a class-token branch and an attention-pooling branch; the two representations are concatenated into a dual feature and classified by an MLP that outputs five class logits.}
\label{fig:dual_aggregation_layer}
\end{figure}

After the Transformer encoder, ROSA-TFormer applies a dual aggregation and classification module that combines the class-token representation with an attention-pooled temporal representation. Let $\mathbf{h}_{cls}$ denote the encoded class token and $\mathbf{h}_t$ denote the encoded temporal tokens. Attention pooling computes:
\begin{equation}
\beta_t=\frac{\exp(a(\mathbf{h}_t))}{\sum_{j=1}^{T}\exp(a(\mathbf{h}_j))},\quad
\mathbf{h}_{pool}=\sum_{t=1}^{T}\beta_t\mathbf{h}_t,
\end{equation}
where $a(\cdot)$ is a small neural scoring function. The final classifier receives $[\mathbf{h}_{cls};\mathbf{h}_{pool}]$ and maps it to five class logits through a two-layer MLP with dropout. Fig.~\ref{fig:dual_aggregation_layer} illustrates this simplified dual-readout classification layer. The model is optimized using weighted cross-entropy with label smoothing to mitigate the smaller sample proportion of \textit{P. sylvestris}.

\subsection{Tabular Temporal Features for Tree-Based Baselines}
To ensure a fair comparison with traditional machine learning methods, the same sequence tensor is converted into a tabular feature vector for RF and XGBoost. The code extracts three groups of features: (1) the flattened raw time series, (2) per-channel summary statistics, and (3) first-order temporal difference statistics. Specifically, each sample contains
\begin{equation}
\mathbf{f}_{\text{flat}}\in\mathbb{R}^{T\times12},
\end{equation}
four band-wise statistics $\{\mu,\sigma,\min,\max\}\in\mathbb{R}^{4\times12}$, and two temporal-difference descriptors $\{\mu_{\Delta},\sigma_{\Delta}\}\in\mathbb{R}^{2\times12}$. The resulting feature vector has
\begin{equation}
T\times12+4\times12+2\times12=12(T+6)
\end{equation}
dimensions per sample, corresponding to 216 dimensions for the monthly dataset and 360 dimensions for the half-month dataset. This design gives the tree-based baselines access to both original temporal observations and compact seasonal descriptors, while the Transformer receives the normalized sequence directly.

\subsection{Initial Baseline Models}
\subsubsection{Random Forest}
Random Forest was included in the initial baseline experiment as a traditional ensemble classifier. In the code, RF is configured with 300 trees, unlimited maximum depth, a fixed random seed, parallel CPU execution, and balanced subsample class weighting. This baseline remains useful for rapid prototyping and checking whether hand-crafted temporal descriptors provide sufficient discriminative information.

\subsubsection{XGBoost}
XGBoost was trained on the same tabular temporal features. The model uses 400 estimators, maximum tree depth 6, learning rate 0.05, subsampling ratio 0.9, column sampling ratio 0.9, and multinomial softmax objective. It provides a strong gradient-boosted tree baseline and is useful for evaluating whether sequential deep learning provides benefits beyond tabular nonlinear modeling.

\subsubsection{1D-CNN}
The 1D-CNN baseline contains three temporal convolution blocks with channel dimensions 64, 128, and 128, each followed by batch normalization and ReLU activation. Adaptive average pooling aggregates the temporal axis before a two-layer fully connected classifier with dropout $p=0.3$. It captures local temporal motifs but lacks the global pairwise temporal comparison provided by self-attention.

\subsubsection{Plain Temporal Transformer}
The PlainTransformer baseline is used as an early-fusion temporal Transformer. It projects each complete 12-channel Sentinel-1/2 token into the same 96-dimensional embedding space with a single linear branch, prepends a class token, adds learnable temporal positional embeddings, and applies the same three-layer Transformer encoder. Unlike ROSA-TFormer, it has no separate SAR/optical tokenization, no sensor-aware gate, and no dual aggregation head. It therefore tests whether the proposed sensor-aware design is more useful than a generic early-fusion Transformer under the same input sequence and split.

\subsection{Training and Evaluation}
For both datasets, GEE records were first reconstructed into point-level annual sequences and then split at the point-sequence level. Monthly-data contains 310/55/157 train/validation/test sequences, while HalfMonth-dataBig contains 2373/419/1197 train/validation/test sequences. This split avoids placing different temporal records of the same point into different subsets. ROSA-TFormer was trained with AdamW, a learning rate of $5\times10^{-4}$, weight decay of $10^{-4}$, batch size 32, and a maximum of 80 epochs. Cosine learning-rate annealing, gradient clipping, label smoothing of 0.02, and early stopping with patience 18 were used. To account for the smaller \textit{P. sylvestris} class, the cross-entropy loss uses inverse-frequency class weights normalized by their mean. The best checkpoint was selected by validation accuracy, using validation loss as a tie-breaker when multiple epochs reached the same validation accuracy. Overall accuracy (OA), macro F1, target-class precision/recall/F1, confusion matrices, and spatial split results were used for evaluation. The revision experiments were run with Python 3.11, PyTorch 2.5.1, CUDA 12.4, and an NVIDIA GeForce RTX 3080 Ti Laptop GPU with 16 GB memory. Random seeds were fixed for Python, NumPy, and PyTorch, and cuDNN deterministic mode was enabled.

\begin{table}[!t]
\centering
\caption{Implementation details reflected in the experimental code.}
\label{tab:implementation}
\begin{tabular}{@{}ll@{}}
\toprule
Item & Setting \\
\midrule
Sequence length & 12 monthly or 24 half-month tokens \\
Input channels & 2 Sentinel-1 + 10 Sentinel-2 bands \\
Train/val/test & 310/55/157 and 2373/419/1197 \\
Deep optimizer & AdamW + cosine annealing \\
Epochs / batch size & max 80 / 32 \\
Learning rate & $5\times10^{-4}$ \\
Weight decay & $10^{-4}$ \\
ROSA-TFormer & 3 layers, 4 heads, $d=96$ \\
Feed-forward dim. & 192 \\
Dropout & 0.15 in encoder, 0.3 in head \\
Trainable params. & 262,136 monthly / 263,288 half-month \\
Tree-feature dim. & 216 monthly / 360 half-month \\
\bottomrule
\end{tabular}
\end{table}

\section{Results}
\subsection{Temporal Characteristics of the Classes}
\begin{figure}[!t]
\centering
\includegraphics[width=\linewidth]{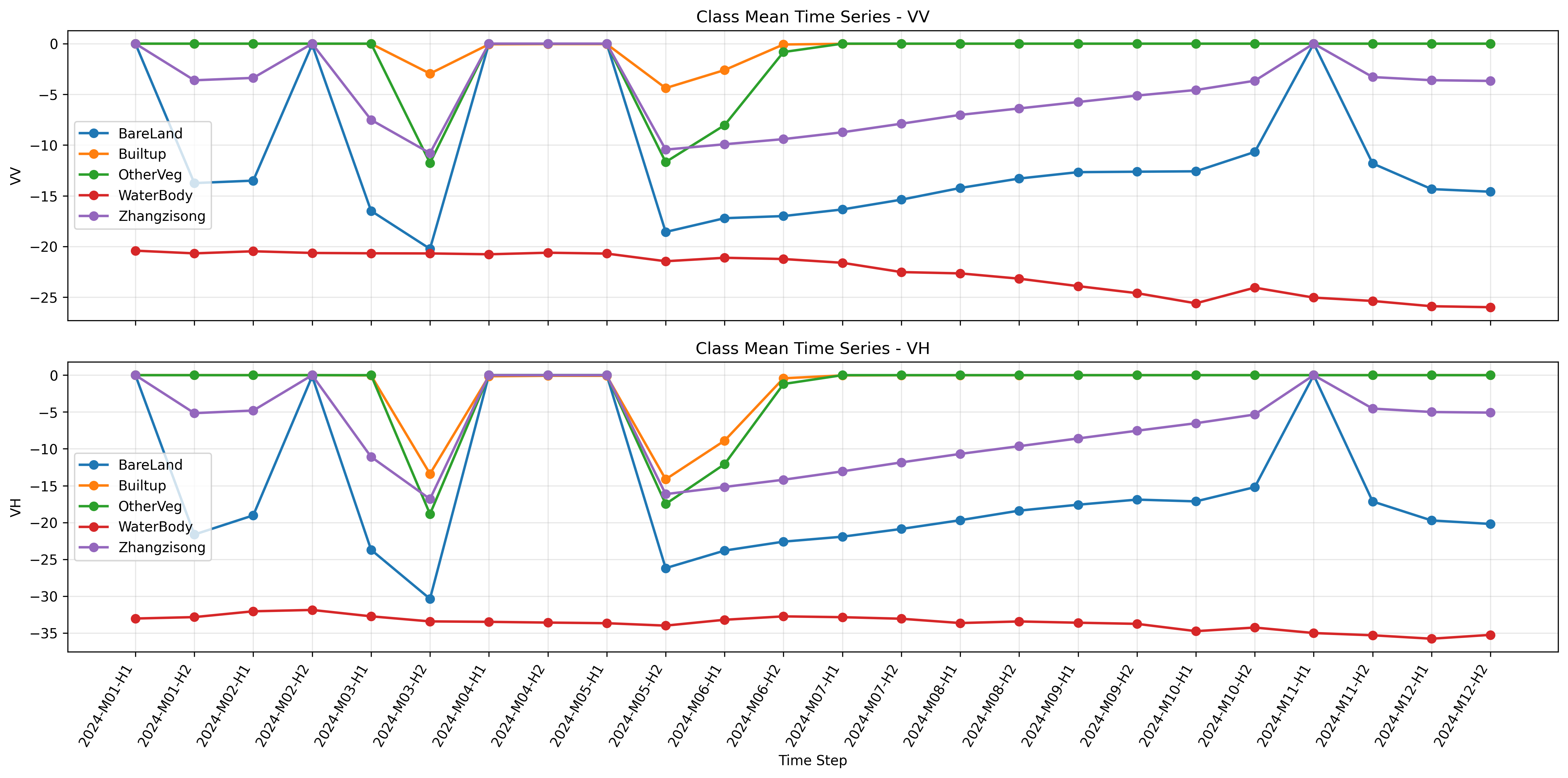}
\caption{Class-mean Sentinel-1 VV/VH time series across 24 half-month windows in 2024.}
\label{fig:s1}
\end{figure}

The annual Sentinel-1 and Sentinel-2 trajectories reveal clear class-level differences. Fig.~\ref{fig:s1} shows that water bodies maintain very low VV and VH backscatter, while built-up areas show high and stable backscatter caused by surface roughness and double-bounce scattering. \textit{P. sylvestris} and other vegetation occupy intermediate ranges, but their temporal stability differs. Evergreen \textit{P. sylvestris} maintains more stable volume scattering during winter than deciduous vegetation.

\begin{figure}[!t]
\centering
\includegraphics[width=\linewidth]{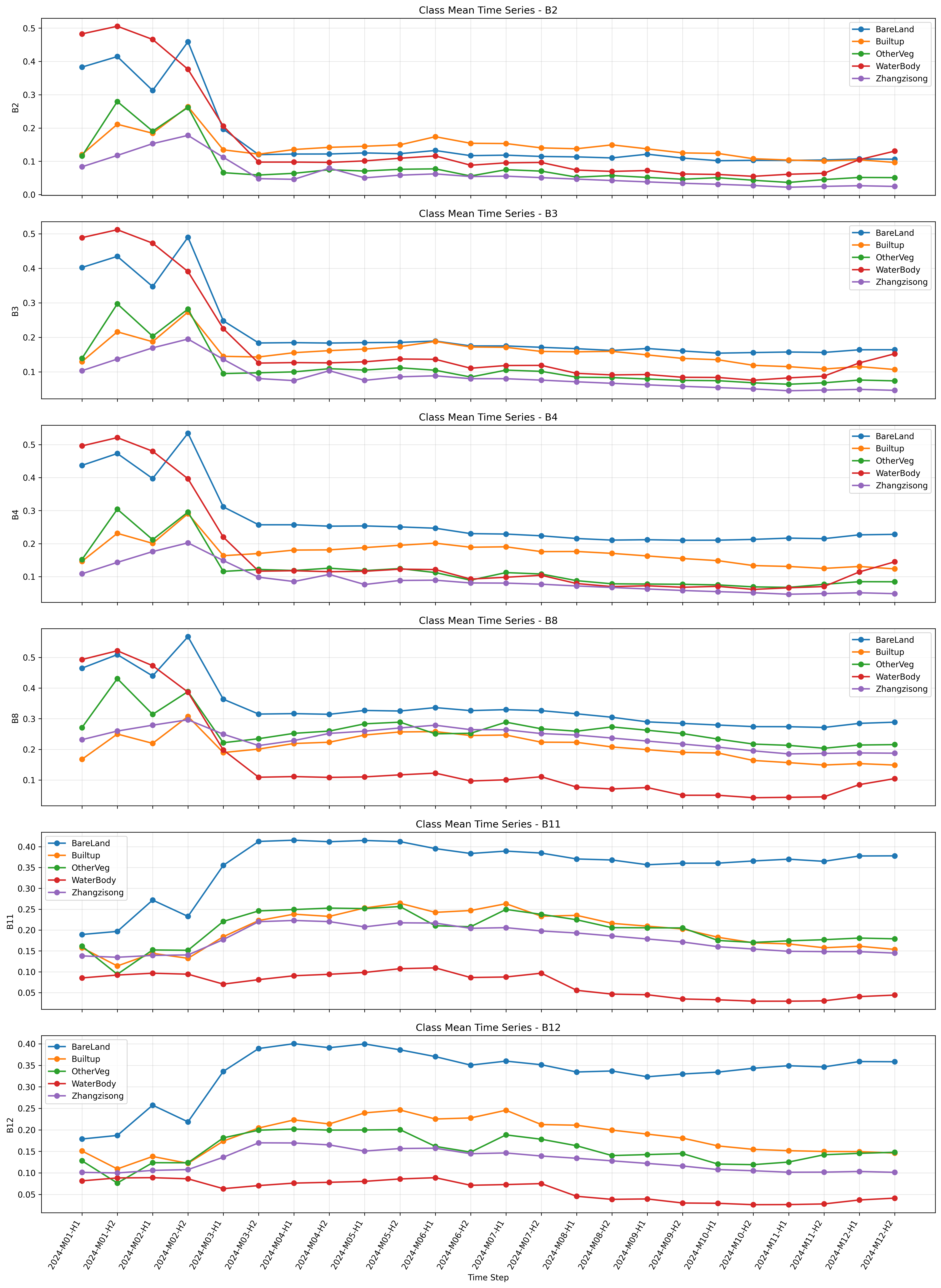}
\caption{Class-mean Sentinel-2 optical time series for selected bands across the same 24 half-month windows.}
\label{fig:s2}
\end{figure}

Fig.~\ref{fig:s2} shows representative Sentinel-2 optical trajectories. The NIR and SWIR bands provide strong evidence for separating vegetation, bare land, and built-up surfaces. \textit{P. sylvestris} preserves a relatively stable winter optical response compared with other vegetation, which shows a stronger leaf-on/leaf-off cycle. This annual contrast is central to the success of the proposed temporal Transformer.

\subsection{Baseline Comparison Under Point-Level Validation}
Table~\ref{tab:baseline_comparison} reports the unified baseline experiment under strict point-level validation. The results show that the classification task is highly separable when complete annual Sentinel-1/2 sequences are available. On Monthly-data, 1D-CNN and the early-fusion Transformer obtain the highest OA and macro F1, whereas ROSA-TFormer maintains perfect target-class F1 but lower OA. On HalfMonth-dataBig, 1D-CNN achieves the highest point-split accuracy, while ROSA-TFormer reaches 99.67\% OA and 99.56\% macro F1, matching XGBoost in OA and slightly exceeding it in macro F1. These results indicate that the proposed model is competitive with strong baselines, but the evidence does not support an over-strong claim that ROSA-TFormer is best on every metric and split.

\begin{table*}[!t]
\centering
\caption{Baseline comparison under strict point-level sequence validation, seed 42. Pine F1 denotes the F1-score of \textit{P. sylvestris} plantation.}
\label{tab:baseline_comparison}
\begin{tabular}{llrrrr}
\toprule
Dataset & Model & Len. & OA & Macro F1 & Pine F1 \\
\midrule
Monthly-data & Random Forest & 12 & 96.82\% & 96.57\% & 94.44\% \\
Monthly-data & XGBoost & 12 & 96.82\% & 96.57\% & 94.44\% \\
Monthly-data & 1D-CNN & 12 & \textbf{98.73\%} & \textbf{98.80\%} & \textbf{100.00\%} \\
Monthly-data & Early-fusion Transformer & 12 & \textbf{98.73\%} & \textbf{98.80\%} & \textbf{100.00\%} \\
Monthly-data & ROSA-TFormer & 12 & 96.82\% & 97.05\% & \textbf{100.00\%} \\
\midrule
HalfMonth-dataBig & Random Forest & 24 & 99.58\% & 99.49\% & 98.91\% \\
HalfMonth-dataBig & XGBoost & 24 & 99.67\% & 99.56\% & 98.91\% \\
HalfMonth-dataBig & 1D-CNN & 24 & \textbf{99.92\%} & \textbf{99.93\%} & \textbf{100.00\%} \\
HalfMonth-dataBig & Early-fusion Transformer & 24 & 99.75\% & 99.64\% & 98.91\% \\
HalfMonth-dataBig & ROSA-TFormer & 24 & 99.67\% & 99.56\% & 98.91\% \\
\bottomrule
\end{tabular}
\end{table*}

\begin{table}[!t]
\centering
\caption{Model size and training cost on HalfMonth-dataBig under the point-level split, seed 42. Tree ensembles are reported by training time only because they do not have trainable neural parameters.}
\label{tab:runtime}
\begin{tabular}{lrrr}
\toprule
Model & Params & Train s & Best epoch \\
\midrule
Random Forest & -- & 1.5 & -- \\
XGBoost & -- & 12.1 & -- \\
1D-CNN & 77,637 & 55.0 & 73 \\
Early-fusion Transformer & 228,965 & 109.1 & 51 \\
ROSA-TFormer & 263,288 & 165.3 & 64 \\
\bottomrule
\end{tabular}
\end{table}

To examine seed sensitivity, the three deep temporal models were additionally evaluated on HalfMonth-dataBig using seeds 42, 7, and 2024. Table~\ref{tab:multiseed} shows that 1D-CNN remains the strongest point-level baseline on average. ROSA-TFormer obtains higher mean OA and macro F1 than the early-fusion Transformer and slightly lower standard deviation, supporting the claim that sensor-aware tokenization provides a competitive and stable Transformer variant.

\begin{table}[!t]
\centering
\caption{Three-seed point-level results on HalfMonth-dataBig, reported as mean $\pm$ standard deviation in percentage points.}
\label{tab:multiseed}
\begin{tabular}{lccc}
\toprule
Model & OA & Macro F1 & Pine F1 \\
\midrule
1D-CNN & \textbf{99.69$\pm$0.46} & \textbf{99.62$\pm$0.60} & \textbf{99.13$\pm$1.51} \\
Early-fusion Transformer & 99.53$\pm$0.26 & 99.42$\pm$0.31 & 98.74$\pm$0.82 \\
ROSA-TFormer & 99.61$\pm$0.17 & 99.49$\pm$0.26 & 98.74$\pm$0.82 \\
\bottomrule
\end{tabular}
\end{table}

\subsection{Per-Class Performance of ROSA-TFormer}
Table~\ref{tab:transformer_report} reports the per-class performance of ROSA-TFormer on the larger HalfMonth-dataBig dataset. Water body is classified perfectly because of its distinctive SAR and optical signals. Bare land, built-up area, and other vegetation also achieve high F1-scores. The target class \textit{P. sylvestris} plantation reaches 100.00\% precision, 97.85\% recall, and 98.91\% F1, confirming that the model captures the evergreen temporal signature of the species while still missing a small number of target samples.

\begin{table}[!t]
\centering
\caption{Per-class classification report of ROSA-TFormer on HalfMonth-dataBig under point-level validation.}
\label{tab:transformer_report}
\begin{tabular}{lcccc}
\toprule
Class & Precision & Recall & F1 & Support \\
\midrule
Bare land & 0.9964 & 0.9964 & 0.9964 & 276 \\
Built-up area & 0.9927 & 0.9963 & 0.9945 & 272 \\
Other vegetation & 0.9964 & 1.0000 & 0.9982 & 275 \\
Water body & 1.0000 & 1.0000 & 1.0000 & 281 \\
\textit{P. sylvestris} plantation & 1.0000 & 0.9785 & 0.9891 & 93 \\
\midrule
Macro average & 0.9971 & 0.9942 & 0.9956 & 1197 \\
\bottomrule
\end{tabular}
\end{table}

The confusion pattern is ecologically interpretable. Only four test samples are misclassified in this split. Two are target-class false negatives: one \textit{P. sylvestris} sample is predicted as bare land, and one is predicted as built-up area. The remaining two errors are one bare-land sample predicted as built-up area and one built-up sample predicted as other vegetation. These errors are consistent with the field context of semi-arid afforestation sites, where sparse plantation canopy, exposed soil, roads, construction land, and rural margins may be spatially adjacent. This observation motivates the future integration of canopy height, topographic position, high-resolution texture, or additional conifer reference classes.

\subsection{Ablation Study}
Table~\ref{tab:ablation} summarizes the ablation experiment on HalfMonth-dataBig. Removing the sensor gate reduces OA from 99.67\% to 99.58\% and decreases the target-class F1 from 98.91\% to 98.38\%, indicating that token-level radar-optical gating provides a measurable benefit. Removing the attention-pooling branch converts the dual aggregation and classification layer into a class-token-only readout and has little effect under the current point split. Removing the class token forces the model to rely only on the attention-pooled temporal representation and slightly lowers OA. The SAR-only Transformer drops sharply to 94.99\% OA, while the optical-only Transformer remains strong at 99.50\% OA. Therefore, Sentinel-2 optical phenology is the dominant signal in this dataset, and Sentinel-1 contributes mainly as complementary structural information within the fused model.

\begin{table}[!t]
\centering
\caption{Ablation study on HalfMonth-dataBig under point-level validation.}
\label{tab:ablation}
\begin{tabular}{lrrr}
\toprule
Model & OA & Macro F1 & Pine F1 \\
\midrule
ROSA-TFormer & \textbf{99.67\%} & \textbf{99.56\%} & 98.91\% \\
No sensor gate & 99.58\% & 99.42\% & 98.38\% \\
No dual aggregation & \textbf{99.67\%} & 99.56\% & 98.91\% \\
No class token & 99.58\% & 99.49\% & 98.91\% \\
No class weighting & \textbf{99.67\%} & 99.56\% & 98.91\% \\
SAR-only Transformer & 94.99\% & 94.67\% & 92.90\% \\
Optical-only Transformer & 99.50\% & 99.42\% & \textbf{98.92\%} \\
\bottomrule
\end{tabular}
\end{table}

\subsection{Sensor Gate and Attention Interpretation}
\begin{figure*}[!t]
\centering
\includegraphics[width=0.92\textwidth]{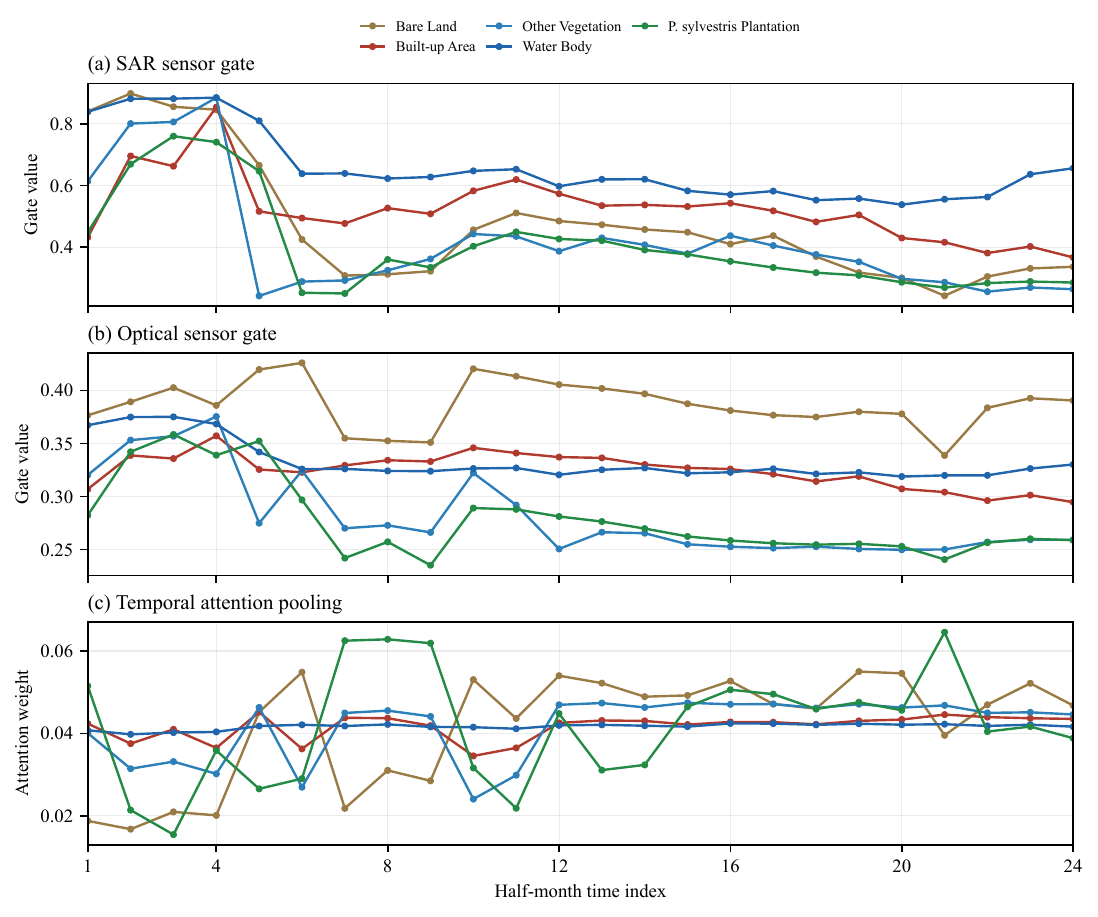}
\caption{Class-mean sensor gate and temporal attention patterns of ROSA-TFormer on HalfMonth-dataBig. Gate values are independent sigmoid responses and therefore do not sum to one.}
\label{fig:gate_attention}
\end{figure*}

Fig.~\ref{fig:gate_attention} visualizes the class-mean sensor gates and temporal attention weights of ROSA-TFormer on the HalfMonth-dataBig test set. The SAR gate is highest on water body on average (0.6569), consistent with the strong separability of water in C-band backscatter. The \textit{P. sylvestris} plantation class shows a lower average SAR gate (0.4028) but a stronger peak temporal attention weight (0.0646), indicating that the model concentrates part of the classification evidence in selected half-month windows rather than using all time steps uniformly. These observations support the interpretation that the sensor gate acts as an adaptive operation on radar-optical token features, while attention pooling provides an explicit temporal readout of discriminative periods.

\subsection{Spatial Split and Temporal Granularity}
To evaluate spatial generalization, a deterministic coordinate-based block split was constructed from the GEE point geometries in HalfMonth-dataBig. This split produced 2217/403/1369 train/validation/test point sequences. Table~\ref{tab:spatial_split} shows that spatial separation changes the ranking of models. The 1D-CNN baseline achieves the highest OA and macro F1, while Random Forest obtains the highest target-class F1. ROSA-TFormer remains competitive, outperforming the early-fusion Transformer in macro F1 and Pine F1 under the same spatial split. This result suggests that the proposed sensor-aware fusion is useful under spatial separation, but it also confirms that the current evidence should be presented as competitive and interpretable rather than universally dominant.

\begin{table}[!t]
\centering
\caption{Spatial block split results on HalfMonth-dataBig.}
\label{tab:spatial_split}
\begin{tabular}{lrrr}
\toprule
Model & OA & Macro F1 & Pine F1 \\
\midrule
Random Forest & 98.90\% & 98.95\% & \textbf{99.39\%} \\
XGBoost & 92.84\% & 91.51\% & 86.32\% \\
1D-CNN & \textbf{99.63\%} & \textbf{99.43\%} & 98.20\% \\
Early-fusion Transformer & 99.42\% & 98.94\% & 96.25\% \\
ROSA-TFormer & 99.34\% & 99.08\% & 97.62\% \\
\bottomrule
\end{tabular}
\end{table}

Finally, the same HalfMonth-dataBig points were aggregated into both 12 monthly tokens and 24 half-month tokens. Table~\ref{tab:granularity} shows that ROSA-TFormer achieved 99.67\% OA in both settings; macro F1 changed only from 99.5635\% to 99.5638\%. Thus, the 24-token sequence provides finer phenological sampling, but its marginal gain is small for the current five-class point-sample setting.

\begin{table}[!t]
\centering
\caption{Same-sample temporal granularity comparison using identical HalfMonth-dataBig points and the same point-level split.}
\label{tab:granularity}
\begin{tabular}{lrrrr}
\toprule
Variant & Len. & Train s & OA & Macro F1 \\
\midrule
Monthly aggregation & 12 & 152.0 & 99.67\% & 99.5635\% \\
Half-month aggregation & 24 & 156.9 & 99.67\% & 99.5638\% \\
\bottomrule
\end{tabular}
\end{table}

\begin{figure}[!t]
\centering
\includegraphics[width=\linewidth]{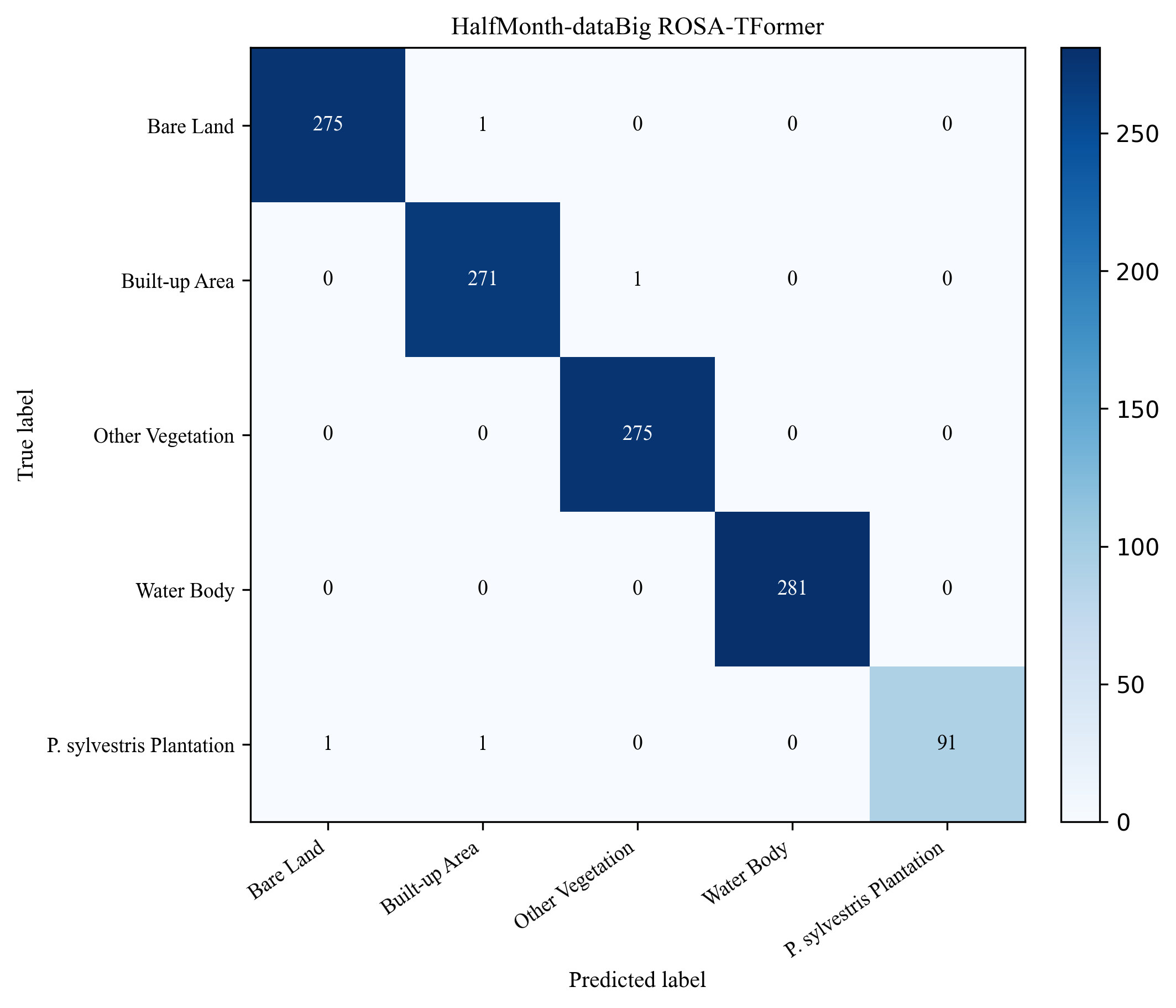}
\caption{Confusion matrix of ROSA-TFormer on HalfMonth-dataBig under strict point-level validation. The new figure uses unified English class names and reports five output classes.}
\label{fig:cm}
\end{figure}

\subsection{Training Behavior}
\begin{figure*}[!t]
\centering
\includegraphics[width=0.86\textwidth]{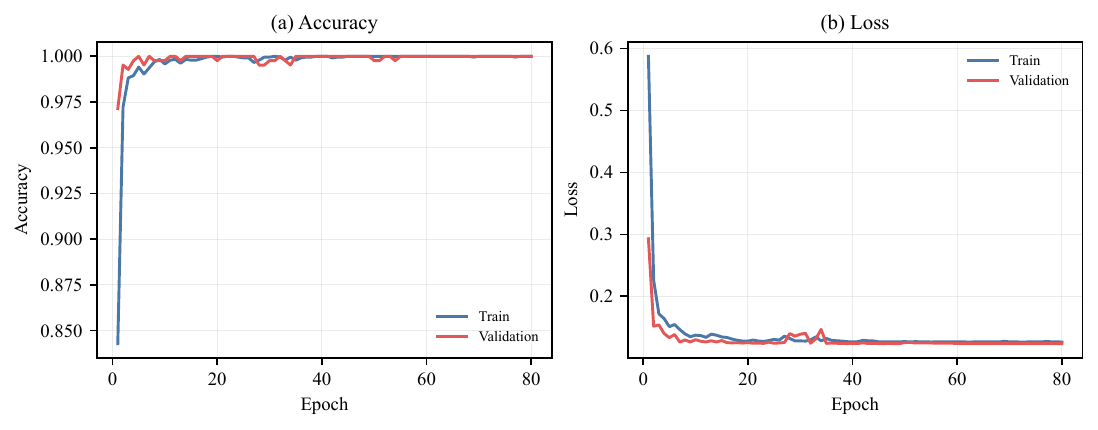}
\caption{Training and validation curves of ROSA-TFormer on HalfMonth-dataBig generated from the unified revision experiment script.}
\label{fig:curves}
\end{figure*}

The training curves in Fig.~\ref{fig:curves} indicate stable convergence of ROSA-TFormer on the larger half-month dataset. Validation accuracy reaches its best value before early stopping, while the use of dropout, label smoothing, and validation-based checkpoint selection reduces overfitting risk.

\section{Discussion}
\subsection{Why Temporal Self-Attention Helps}
The classification of \textit{P. sylvestris} depends on annual temporal structure. During winter, evergreen conifers retain canopy signals when deciduous vegetation and cropland decline. During the growing season, optical differences may become less obvious because many vegetation classes reach high greenness. SAR observations add complementary information by responding to canopy structure and moisture. The proposed Transformer can integrate all of these cues by allowing each time step to attend to every other time step. This is a natural fit for full-year vegetation analysis.

By contrast, tabular models require explicit manual construction of temporal statistics, and convolutional models emphasize local neighborhoods. The baseline results show that these alternatives are strong in the present five-class point-sample task, especially the 1D-CNN under point-level validation. Across three seeds, ROSA-TFormer does not surpass 1D-CNN, but it achieves higher mean OA and macro F1 than the early-fusion Transformer with slightly lower variability. Therefore, the contribution of ROSA-TFormer should not be interpreted only as a universal accuracy gain. Its value lies in a sensor-aware temporal representation that keeps SAR and optical embeddings physically distinguishable, exposes gate and attention weights for interpretation, and remains competitive under both point-level and spatial block validation.

The implementation also clarifies why the comparison is meaningful. RF and XGBoost are not weak baselines: they receive flattened observations and seasonal descriptors from the same annual sequence, forming a 216-dimensional vector for monthly data or a 360-dimensional vector for half-month data. The deep models receive normalized sequences directly. The resulting comparison therefore evaluates different representation strategies for the same GEE-derived annual observations rather than comparing models with different input information.

\subsection{Monthly Versus Half-Month Temporal Granularity}
The comparison between Monthly-data and HalfMonth-dataBig is influenced by both temporal granularity and sample size. Therefore, an additional same-sample experiment was conducted by aggregating the dataBig points from the northern Shaanxi study area into 12 monthly tokens and comparing them with the original 24 half-month tokens. ROSA-TFormer achieved the same OA (99.67\%) in both settings, and macro F1 changed only marginally from 99.5635\% to 99.5638\%. This indicates that 24-token compositing provides finer phenological detail, but the marginal gain is small for the present five-class point-sample task.

For future map production, the half-month dataset may still be preferable when GEE data availability and computing resources allow dense compositing, because it preserves more detailed phenological transitions and may be more useful in areas with cloud gaps, short phenological events, or more similar conifer classes. The monthly representation is a lighter alternative when export limits or computational cost make dense temporal sampling less convenient. This conclusion currently refers to point-level classification evidence; wall-to-wall mapping still requires raster-level inference and independent map validation.

\subsection{Implications for Shelterbelt Monitoring}
The Three-North Shelter Forest Program is a long-term ecological engineering project, and northern Shaanxi is one of the important semi-arid landscapes where plantation monitoring must distinguish planted conifers from heterogeneous background classes. Monitoring this region requires not only detecting forest cover but also identifying key planted species and evaluating their spatial distribution. The proposed method contributes to this need by distinguishing \textit{P. sylvestris} from classes that frequently coexist in northern Shaanxi shelterbelt landscapes. This is more operationally meaningful than a simplified forest/non-forest classification.

The use of freely available Sentinel-1 and Sentinel-2 data on GEE also supports repeatable monitoring in northern Shaanxi. Half-month compositing balances temporal detail and robustness to cloud contamination, while the cloud-computing workflow reduces the burden of downloading and preprocessing large quantities of imagery. In practice, the framework could be extended to annual map updating, plantation change screening, and field inspection planning once wall-to-wall inference and independent validation are added.

\subsection{Why Point-Level Validation Matters}
The optimized experiment uses a point-level split rather than a raw-record split. This design is important because monthly or half-month records from the same point are temporally correlated. If different records from one point appear in both training and testing subsets, the reported accuracy can overestimate the ability of a model to generalize to unseen plantation sites. Reconstructing complete annual sequences before splitting is therefore a stricter and more appropriate protocol for evaluating a temporal model.

\subsection{Limitations}
Several limitations remain. First, the current samples are point-level sequences from the northern Shaanxi study area and adjacent shelterbelt landscapes. Wall-to-wall mapping requires applying the model to image grids and validating the resulting map with independent field or visual interpretation data. Second, only five classes are considered. In future work, \textit{P. sylvestris} should be separated from other conifers such as \textit{Pinus tabuliformis}, larch, and cypress where reference data are available. Third, the strongest point-level baseline is 1D-CNN in the three-seed experiment, and the deterministic spatial split also shows that different metrics favor different models. This means that the current evidence supports ROSA-TFormer as a competitive and interpretable radar-optical temporal model rather than as a universally dominant classifier. Fourth, the model uses only spectral and SAR time series; high-resolution texture, topography, climate variables, or canopy height data could further improve discrimination. Fifth, the local revision pipeline starts from GEE-exported point CSV files. Before formal submission, the exact GEE export script and metadata for scene filtering, cloud masking, and compositing should be archived as supplementary material to improve full reproducibility.

\section{Conclusion}
This paper has presented ROSA-TFormer, a Radar-Optical Sensor-Aware Temporal Transformer for \textit{Pinus sylvestris} var. \textit{mongolica} plantation classification in northern Shaanxi using GEE-derived Sentinel-1/2 monthly and half-month time series. The study was motivated by the practical need to monitor artificial plantation forests and shelterbelt plantations under the Three-North Shelter Forest Program. By organizing VV/VH SAR backscatter and ten optical bands into annual temporal sequences, the model learns the evergreen phenological and structural characteristics of \textit{P. sylvestris}.

Under strict point-level sequence validation, ROSA-TFormer achieved 99.67\% overall accuracy, 99.56\% macro F1, and 98.91\% \textit{P. sylvestris} F1 on HalfMonth-dataBig. Across three seeds, it averaged 99.61\% overall accuracy and 99.49\% macro F1, slightly outperforming the early-fusion Transformer on average but remaining below 1D-CNN. In the deterministic coordinate-based spatial block split, ROSA-TFormer achieved 99.34\% overall accuracy and 99.08\% macro F1, remaining competitive with strong deep-learning baselines while outperforming the early-fusion Transformer in macro F1 and target-class F1. The ablation experiment showed that the sensor-aware gate improves both macro F1 and target-class F1 compared with a no-gate variant. These results indicate that radar-optical sensor-aware temporal self-attention is a promising and interpretable tool for species-level shelterbelt classification and provides a point-level foundation for future northern Shaanxi \textit{P. sylvestris} mapping workflows.

\section*{Acknowledgment}
The author thanks the European Space Agency for free access to Sentinel-1 and Sentinel-2 data and Google Earth Engine for providing scalable cloud-based geospatial data access and sampling. The manuscript was developed from project code, data tables, model reports, and experimental figures generated in the current research workflow.

\balance


\begin{thebibliography}{99}

\bibitem{zheng2012}
X. Zheng, J. J. Zhu, Q. L. Yan, and L. N. Song, ``Effects of land use changes on the groundwater table and the decline of \textit{Pinus sylvestris} var. \textit{mongolica} plantations in southern Horqin Sandy Land, Northeast China,'' \emph{Agricultural Water Management}, vol. 109, pp. 94--106, 2012.

\bibitem{zhai2023}
J. Zhai, Y. Liu, J. Huang, X. Liu, X. Xu, Z. Wang, and J. Fang, ``Assessing the effects of China's Three-North Shelter Forest Program over 40 years,'' \emph{Science of The Total Environment}, vol. 857, p. 159354, 2023.

\bibitem{li2024trees}
P. Liu, C. Ren, Z. Wang, M. Jia, W. Yu, H. Ren, and C. Xia, ``Evaluating the potential of Sentinel-2 time series imagery and machine learning for tree species classification in a mountainous forest,'' \emph{Remote Sensing}, vol. 16, no. 2, p. 293, 2024.

\bibitem{hemmerling2021}
J. Hemmerling, M. Pflugmacher, S. Hostert, and T. Frantz, ``Mapping temperate forest tree species using dense Sentinel-2 time series,'' \emph{Remote Sensing of Environment}, vol. 267, p. 112743, 2021.

\bibitem{gorelick2017}
N. Gorelick, M. Hancher, M. Dixon, S. Ilyushchenko, D. Thau, and R. Moore, ``Google Earth Engine: Planetary-scale geospatial analysis for everyone,'' \emph{Remote Sensing of Environment}, vol. 202, pp. 18--27, 2017.

\bibitem{torres2012}
R. Torres, P. Snoeij, D. Geudtner, D. Bibby, M. Davidson, E. Attema, P. Potin, B. Rommen, N. Floury, M. Brown, I. N. Traver, P. Deghaye, B. Duesmann, B. Rosich, N. Miranda, C. Bruno, M. L'Abbate, R. Croci, A. Pietropaolo, M. Huchler, and F. Rostan, ``GMES Sentinel-1 mission,'' \emph{Remote Sensing of Environment}, vol. 120, pp. 9--24, 2012.

\bibitem{drusch2012}
M. Drusch, U. Del Bello, S. Carlier, O. Colin, V. Fernandez, F. Gascon, B. Hoersch, C. Isola, P. Laberinti, P. Martimort, A. Meygret, F. Spoto, O. Sy, F. Marchese, and P. Bargellini, ``Sentinel-2: ESA's optical high-resolution mission for GMES operational services,'' \emph{Remote Sensing of Environment}, vol. 120, pp. 25--36, 2012.

\bibitem{breiman2001}
L. Breiman, ``Random forests,'' \emph{Machine Learning}, vol. 45, no. 1, pp. 5--32, 2001.

\bibitem{belgiu2016}
M. Belgiu and L. Dr\u{a}gu\c{t}, ``Random forest in remote sensing: A review of applications and future directions,'' \emph{ISPRS Journal of Photogrammetry and Remote Sensing}, vol. 114, pp. 24--31, 2016.

\bibitem{chen2016}
T. Chen and C. Guestrin, ``XGBoost: A scalable tree boosting system,'' in \emph{Proc. ACM SIGKDD Int. Conf. Knowledge Discovery and Data Mining}, 2016, pp. 785--794.

\bibitem{gomez2016}
C. Gomez, J. C. White, and M. A. Wulder, ``Optical remotely sensed time series data for land cover classification: A review,'' \emph{ISPRS Journal of Photogrammetry and Remote Sensing}, vol. 116, pp. 55--72, 2016.

\bibitem{belgiu2018}
M. Belgiu and O. Csillik, ``Sentinel-2 cropland mapping using pixel-based and object-based time-weighted dynamic time warping analysis,'' \emph{Remote Sensing of Environment}, vol. 204, pp. 509--523, 2018.

\bibitem{dostalova2018}
A. Dost\'{a}lov\'{a}, W. Wagner, M. Milenkovi\'{c}, and M. Hollaus, ``Annual seasonality in Sentinel-1 signal for forest mapping and forest type classification,'' \emph{International Journal of Remote Sensing}, vol. 39, no. 21, pp. 7738--7760, 2018.

\bibitem{clerici2017}
N. Clerici, C. A. Valbuena Calder\'{o}n, and J. M. Posada, ``Fusion of Sentinel-1A and Sentinel-2A data for land cover mapping: A case study in the lower Magdalena region, Colombia,'' \emph{Journal of Maps}, vol. 13, no. 2, pp. 718--726, 2017.

\bibitem{tavares2019}
P. A. Tavares, N. E. S. Beltr\~{a}o, U. S. Guimar\~{a}es, and A. C. Teodoro, ``Integration of Sentinel-1 and Sentinel-2 for classification of LULC in urban areas combining spectral and textural information,'' \emph{Remote Sensing}, vol. 11, no. 12, p. 1470, 2019.

\bibitem{zhong2019}
L. Zhong, L. Hu, and H. Zhou, ``Deep learning based multi-temporal crop classification,'' \emph{Remote Sensing of Environment}, vol. 221, pp. 430--443, 2019.

\bibitem{vaswani2017}
A. Vaswani, N. Shazeer, N. Parmar, J. Uszkoreit, L. Jones, A. N. Gomez, L. Kaiser, and I. Polosukhin, ``Attention is all you need,'' in \emph{Advances in Neural Information Processing Systems}, 2017, pp. 5998--6008.

\bibitem{russwurm2020}
M. Ru\ss{}wurm and M. K\"{o}rner, ``Self-attention for raw optical satellite time series classification,'' \emph{ISPRS Journal of Photogrammetry and Remote Sensing}, vol. 169, pp. 421--435, 2020.

\bibitem{russwurm2019}
M. Ru\ss{}wurm and M. K\"{o}rner, ``Multi-temporal land cover classification with sequential recurrent encoders,'' \emph{ISPRS International Journal of Geo-Information}, vol. 7, no. 4, p. 129, 2018.

\bibitem{yuan2023}
Y. Yuan, L. Lin, Q. Liu, R. Hang, and Z. Zhou, ``Self-supervised pre-training for remote sensing time series classification,'' \emph{IEEE Transactions on Geoscience and Remote Sensing}, vol. 61, pp. 1--15, 2023.

\bibitem{gao1996}
B. C. Gao, ``NDWI: A normalized difference water index for remote sensing of vegetation liquid water from space,'' \emph{Remote Sensing of Environment}, vol. 58, no. 3, pp. 257--266, 1996.

\bibitem{huete2002}
A. Huete, K. Didan, T. Miura, E. P. Rodriguez, X. Gao, and L. G. Ferreira, ``Overview of the radiometric and biophysical performance of the MODIS vegetation indices,'' \emph{Remote Sensing of Environment}, vol. 83, no. 1--2, pp. 195--213, 2002.

\bibitem{key2006}
C. H. Key and N. C. Benson, ``Landscape assessment: ground measure of severity, the Composite Burn Index,'' in \emph{FIREMON: Fire Effects Monitoring and Inventory System}, USDA Forest Service, RMRS-GTR-164-CD, 2006.

\bibitem{kim2012}
Y. Kim, T. Jackson, R. Bindlish, H. Lee, and S. Hong, ``Radar vegetation index for estimating the vegetation water content of rice and soybean,'' \emph{IEEE Geoscience and Remote Sensing Letters}, vol. 9, no. 4, pp. 564--568, 2012.

\bibitem{zhu2008}
J. Zhu, F. Li, M. Xu, H. Kang, and X. Wu, ``The role of ectomycorrhizal fungi in alleviating pine decline in semiarid sandy soil of northern China: An experimental approach,'' \emph{Annals of Forest Science}, vol. 65, no. 3, p. 304, 2008.

\bibitem{mutanga2016}
O. Mutanga, T. Dube, and F. Ahmed, ``Progress in remote sensing of grass senescence: A review,'' \emph{Geocarto International}, vol. 31, no. 6, pp. 621--635, 2016.

\end{thebibliography}
\end{document}